# Neurosymbolic Artificial Intelligence (NSAI) based Algorithm for predicting the Impact Strength of Additive Manufactured Polylactic Acid (PLA) Specimens

Akshansh Mishra[1,*], Vijaykumar S Jatti[2]

[1]School of Industrial and Information Engineering, Politecnico Di Milano, Milan, Italy

[2] Department of Mechanical Engineering, Symbiosis Institute of Technology, Pune, India

**Abstract:**

In this study, we introduce application of Neurosymbolic Artificial Intelligence (NSAI) for predicting the impact strength of additive manufactured polylactic acid (PLA) components, representing the first-ever use of NSAI in the domain of additive manufacturing. The NSAI model amalgamates the advantages of neural networks and symbolic AI, offering a more robust and accurate prediction than traditional machine learning techniques. Experimental data was collected and synthetically augmented to 1000 data points, enhancing the model's precision. The Neurosymbolic model was developed using a neural network architecture comprising input, two hidden layers, and an output layer, followed by a decision tree regressor representing the symbolic component. The model's performance was benchmarked against a Simple Artificial Neural Network (ANN) model by assessing mean squared error (MSE) and R-squared ($R^2$) values for both training and validation datasets.

The results reveal that the Neurosymbolic model surpasses the Simple ANN model, attaining lower MSE and higher $R^2$ values for both training and validation sets. This innovative application of the Neurosymbolic approach in estimating the impact strength of additive manufactured PLA components underscores its potential for optimizing the additive manufacturing process. Future research could investigate further refinements to the Neurosymbolic model, extend its application to other materials and additive manufacturing processes, and incorporate real-time monitoring and control for enhanced process optimization.

1. **Introduction**

Additive manufacturing refers to the fabrication technique where an object is constructed by sequentially depositing material in layers. This approach contrasts with subtractive manufacturing, which involves carving out a desired shape from a solid block of material. Although the term "additive manufacturing" can encompass any process in which an object is formed by the accumulation of material, it is predominantly associated with 3D printing [1-4]. The technology first emerged in the 1980s as a method for rapid prototyping, enabling the quick production of non-functional models without the conventional time-consuming and costly procedures associated with prototype development [5-7]. As additive manufacturing technologies evolved, their applications expanded to include rapid tooling, which facilitated the creation of molds for end-use products. By the early 21st century, additive manufacturing techniques were being employed for the fabrication of functional components. In recent years, industry giants such as Boeing and General Electric have adopted additive manufacturing as a crucial component of their production processes.

Traditional machining processes, such as milling, drilling, rolling, and forming, continue to dominate medium to large-scale production. However, over the past decade, additive manufacturing (AM) has emerged as a disruptive force, offering new opportunities for customized, small-scale production. Conventional manufacturing techniques struggle to achieve optimized production, whereas additive manufacturing can easily facilitate this with minimal tooling changes and significantly reduced manufacturing time. Despite its advantages, additive manufacturing presents unique challenges depending on the specific technology employed, including vat polymerization, powder bed fusion, material extrusion, material jetting, binder jetting, direct energy deposition, or sheet lamination [8-10]. Advances in post-processing techniques have significantly enhanced additive manufacturing capabilities, transforming it from a prototyping-focused approach to a viable method for producing finished products.

Many organizations view digitization and automation as critical factors for advancing additive manufacturing. Consequently, an increasing number of manufacturers are adopting cloud-based solutions and incorporating various algorithms into their 3D printing systems to fully harness the technology's potential. As a digital process, 3D printing is an integral component of Industry 4.0, an era characterized by the growing use of artificial intelligence (AI), such as machine learning, to optimize the value chain [11-15]. AI has the capacity to rapidly process vast amounts of complex data, making it increasingly valuable for decision-making. Machine learning, a subset of AI, refers to systems or software that employ algorithms to analyze data and subsequently identify patterns or derive solutions. While some may believe that machine learning is a recent development, its origins can be traced back to the 1940s, when researchers began emulating brain neurons using electrical circuits. In 1957, the Mark I Perceptron marked a significant milestone in the field, as the machine was capable of independently classifying input data. By learning from past errors, the device continuously improved its classification capabilities. This early success laid the groundwork for ongoing research, as scientists became captivated by the technology's possibilities and potential. Today, AI is encountered daily across various aspects of life, from speech recognition and intelligent chatbots to personalized treatment plans. Machine learning continues to be employed in a wide array of applications.

## 2. Problem Statement

In recent years, additive manufacturing techniques have gained prominence due to their ability to create complex and customized structures, particularly with the growing demand for sustainable materials like Polylactic Acid (PLA). However, accurately predicting the impact strength of additive manufactured PLA components remains a challenge, which directly affects their performance and application potential. This research paper aims to address this issue by employing two distinct predictive models: a Neurosymbolic-based algorithm and a simple Artificial Neural Network (ANN) model. The problem statement for this research work is to investigate the efficacy of these two approaches in accurately predicting the impact strength of additive manufactured PLA components and to determine which model provides better prediction performance and insights for practical applications.

## 3. Concept of Neurosymbolic Artificial Intelligence (NSAI)

Neuro-symbolic artificial intelligence represents an emerging field in AI research, aiming to integrate the advantages of traditional rule-based AI methodologies with contemporary deep learning techniques [16]. Symbolic models offer several benefits, including the requirement for few input samples, effective generalization to new problems, and a conceptually straightforward internal functionality when compared to deep learning models. However, these models necessitate considerable manual tuning, making them challenging to develop for complex problems.

Neuro-symbolic AI strives to harness the strengths of both deep learning and symbolic approaches. Deep learning has demonstrated remarkable success in extracting intricate features from data for tasks such as object detection and natural language processing. In contrast, symbolic AI excels at formalizing human-like reasoning processes. The primary goal of neuro-symbolic AI is to employ deep learning techniques to extract features from data and then manipulate these features using symbolic methodologies, thus capitalizing on the best aspects of both fields.

The framework implemented in the present work is shown in Figure 1.

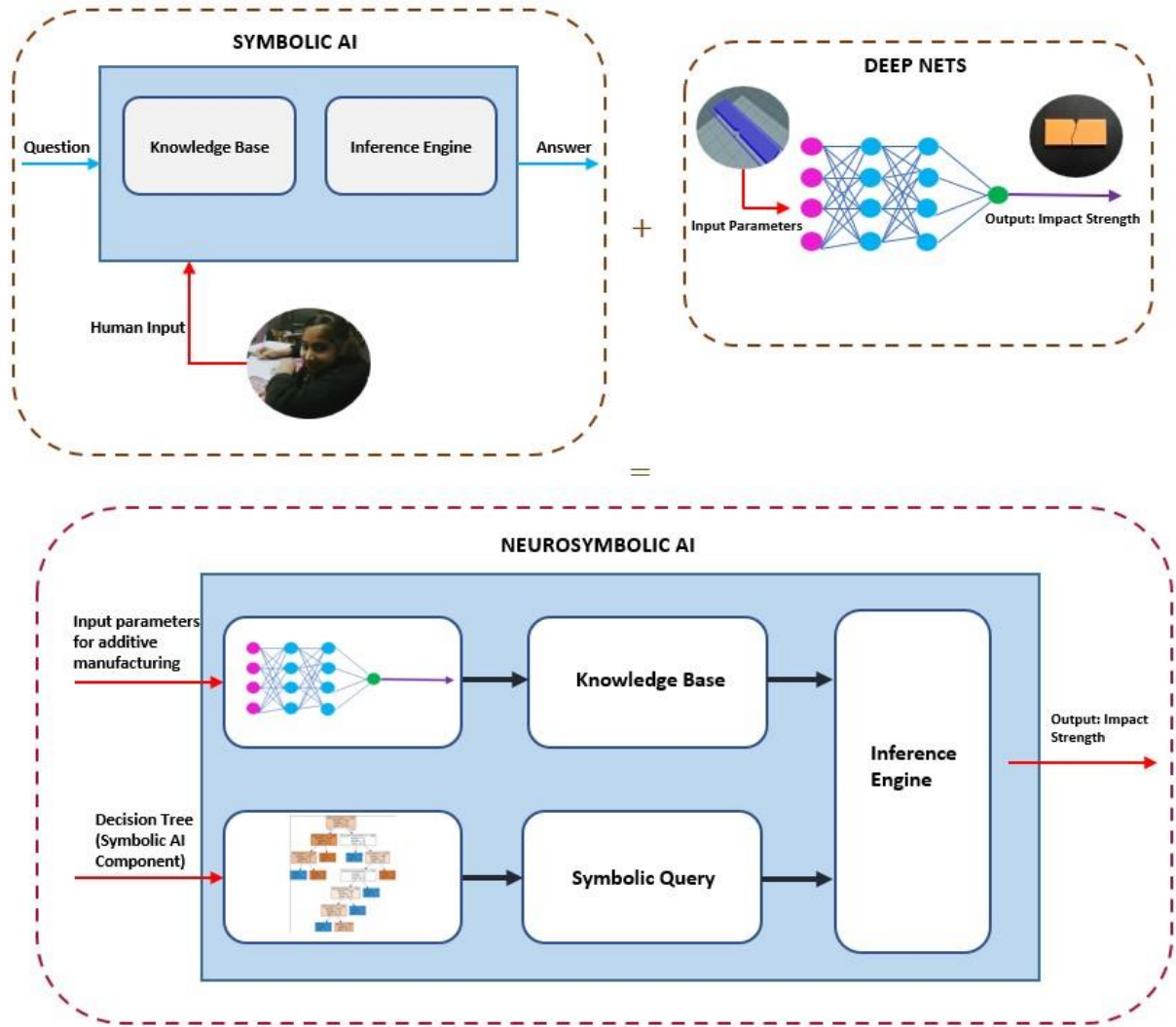

Figure 1. Framework of NSAI implemented in the present work

### 4. Need of Synthetic Data Generation in Additive Manufacturing

Synthetic data generation offers significant advantages in additive manufacturing (AM) by expanding available data, enabling design exploration, reducing costs, and maintaining data privacy. By supplementing existing datasets with artificially generated data, synthetic data generation can improve the accuracy and dependability of machine learning models and computational tools used for process optimization, defect detection, and material property prediction. In addition, synthetic data can facilitate the exploration of various design parameters and their effects on part performance, material properties, and manufacturing efficiency. This helps to optimize designs and reduce the reliance on costly and time-consuming physical prototyping. Synthetic data generation can also lower experimental costs by decreasing the number of physical experiments required, which is particularly beneficial in AM due to the high expenses associated with materials and equipment usage.

Figure 2 presents synthetic data generated using a sine function, augmented with random noise to emulate real-world observations. The original data points are depicted as blue circular markers, while the magenta diamond markers represent the synthetic data points. This illustration highlights the capability of synthetic data generation to produce additional data points that maintain the fundamental pattern of the original data, while also introducing some variability.

Synthetic data generation primarily involves analyzing the given data to discern its distribution, correlations, and underlying patterns. Subsequently, new data points are generated based on this knowledge, frequently employing statistical models, sampling techniques, or machine learning algorithms. The objective is to produce synthetic data that mirrors the original data but includes added variation, ultimately enhancing the robustness of machine learning models and other analytical processes.

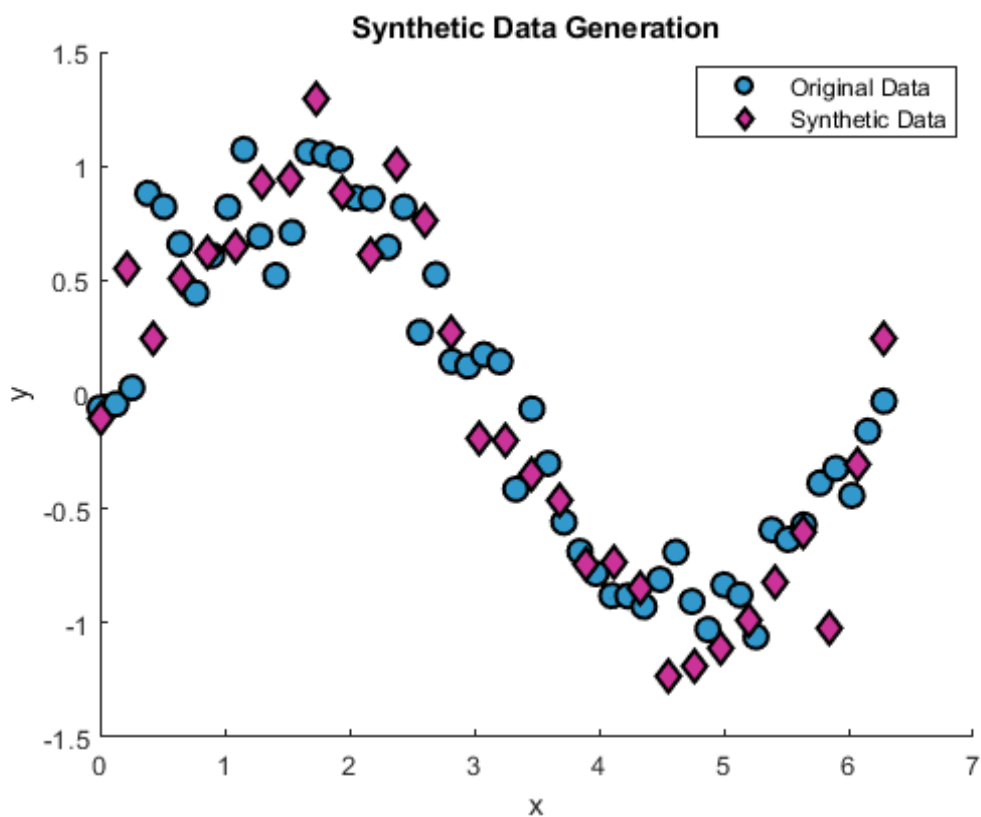

Figure 2. Visualizing Original Data and generated Synthetic data

### 5. Material and Methods

Figure 3 showcases the Fused Deposition Modeling (FDM) process, which entails constructing three-dimensional objects in a layer-by-layer manner using thermoplastic materials like polylactic acid (PLA). The procedure commences with a computer-aided design (CAD) model that is transformed into a suitable file format and divided into thin horizontal layers with the help of dedicated software. These layers produce a set of instructions, or G-code, for the 3D printer to execute. The printer's extruder warms the PLA filament and deposits it through a nozzle onto the build platform, constructing the object one layer at a time. The PLA material merges with the preceding layer and solidifies upon cooling, culminating in the completed 3D object. Support structures may be necessary for intricate geometries or overhangs, and post-processing methods such as sanding or painting can be employed for final touches.

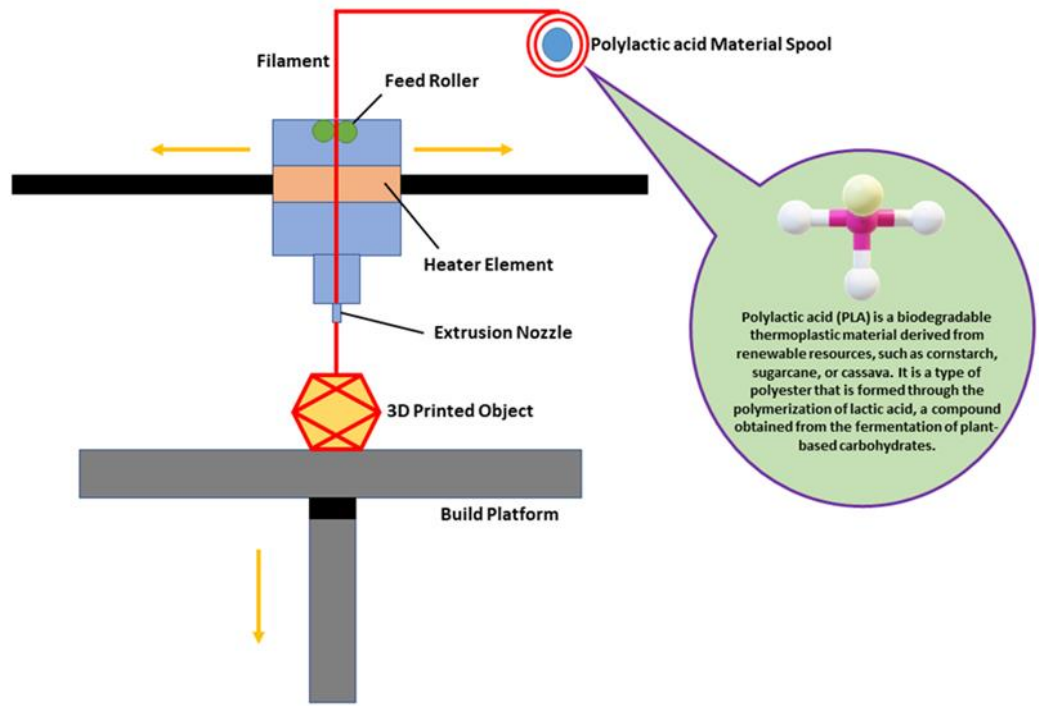

Figure 3. Schematic representation of FDM Process

The FDM process was utilized to fabricate impact strength specimens in compliance with ASTM D256 standard specifications, employing a Creality Ender 3 machine with a bed size of 220 x 220 x 250 mm, as depicted in Figure 4. The component design was developed using CATIA software and transformed into an STL file. Subsequently, the file was converted into a machine-interpretable G-code file through the Cura engine within Repetier software, as demonstrated in Figure 5.

Table 1 presents the input and output parameters of the experimental work. The resulting experimental data is converted into a CSV format file and imported into the Google Colab platform for application of the neurosymbolic programming algorithm. To enhance the model's accuracy, the available data is synthetically expanded to 1,000 data points.

The neurosymbolic programming approach merges the advantages of neural networks and symbolic AI. The neural network structure consists of a series of densely connected layers, with 32 and 16 hidden units and a single output neuron. A Rectified Linear Unit (ReLU) activation function introduces nonlinearity to the model. The network is trained using the Adam optimizer and the mean squared error loss function. Training and validation datasets are employed to fit the model for 2,000 epochs, with a batch size of 32. The resulting trained neural network model is then utilized to extract learned features from the input data. A decision tree, functioning as the symbolic component of the model, is created using the DecisionTreeRegressor from the scikit-learn library. The decision tree's maximum depth is limited to four to prevent overfitting. The model's performance is evaluated by predicting output values for the training and validation sets, using the learned features as inputs. Mean squared error (MSE) and R-squared ($R^2$) values are calculated to assess the model's performance. The obtained metric features are subsequently compared to a simple Artificial Neural Network (ANN) model.

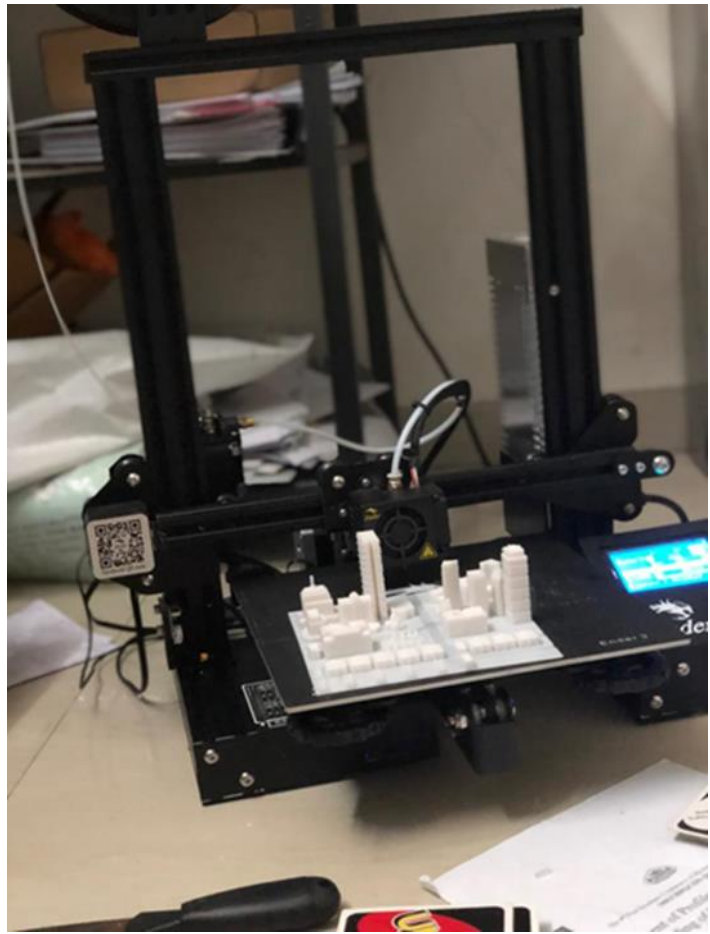

Figure 4. Ender 3 3D printer

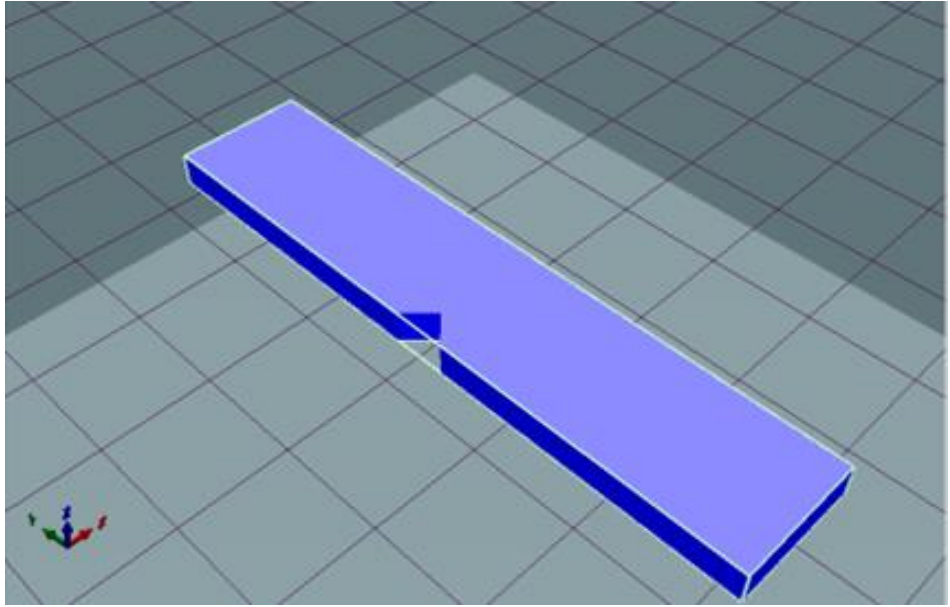

a)

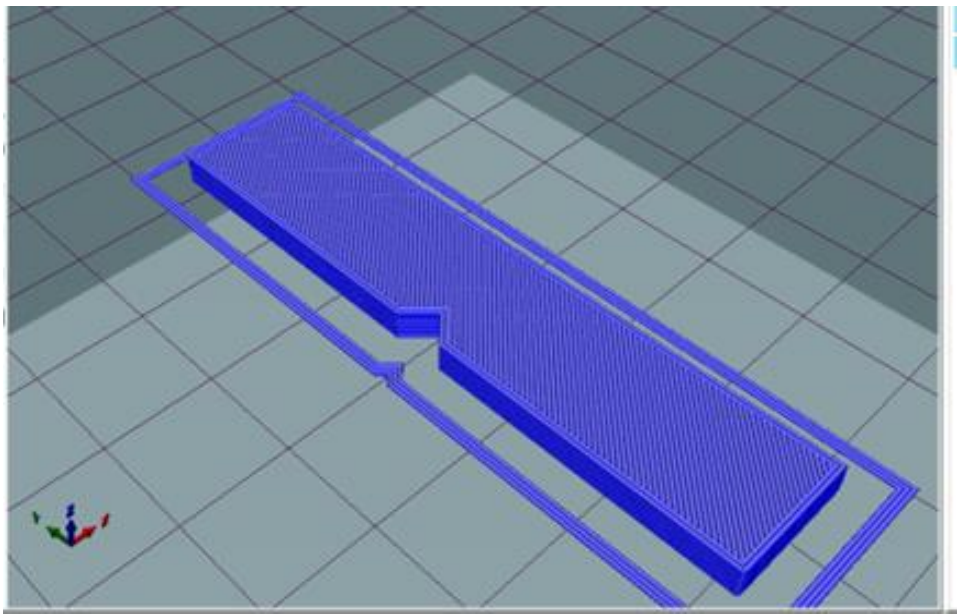

b)

Figure 5. Impact Test specimen a) Before slicing , b) After slicing

Table 1. Experimental Dataset

| Infill percentage (%) | Layer height (mm) | Print speed (mm/sec) | Extrusion temperature (°C) | Impact strength (kJ/m$^2$) |
|---|---|---|---|---|
| 78 | 0.32 | 35 | 220 | 1.55 |
| 10.5 | 0.24 | 50 | 210 | 1.59 |
| 33 | 0.16 | 35 | 220 | 3.2 |
| 33 | 0.32 | 35 | 200 | 3.32 |
| 33 | 0.16 | 65 | 200 | 3.31 |
| 100 | 0.24 | 50 | 210 | 3.37 |
| 78 | 0.16 | 35 | 200 | 3.31 |
| 33 | 0.32 | 65 | 200 | 3.25 |
| 78 | 0.32 | 65 | 200 | 3.31 |
| 33 | 0.16 | 65 | 220 | 3.27 |
| 78 | 0.16 | 35 | 220 | 3.35 |
| 55.5 | 0.24 | 50 | 210 | 3.22 |
| 33 | 0.32 | 35 | 220 | 3.3 |
| 55.5 | 0.24 | 50 | 190 | 3.37 |
| 55.5 | 0.24 | 50 | 210 | 3.38 |
| 78 | 0.32 | 65 | 220 | 3.2 |
| 55.5 | 0.24 | 50 | 210 | 3.38 |
| 55.5 | 0.24 | 50 | 210 | 3.36 |
| 55.5 | 0.24 | 50 | 230 | 1.71 |
| 33 | 0.32 | 65 | 220 | 3.32 |
| 55.5 | 0.24 | 50 | 210 | 3.47 |
| 55.5 | 0.24 | 80 | 210 | 3.38 |
| 78 | 0.16 | 65 | 200 | 3.35 |
| 55.5 | 0.24 | 20 | 210 | 3.52 |
| 55.5 | 0.08 | 50 | 210 | 3.37 |
| 55.5 | 0.4 | 50 | 210 | 3.17 |
| 55.5 | 0.24 | 50 | 210 | 3.47 |
| 78 | 0.32 | 35 | 200 | 3.41 |
| 55.5 | 0.24 | 50 | 210 | 3.43 |
| 78 | 0.16 | 65 | 220 | 3.2 |
| 33 | 0.16 | 35 | 200 | 3.35 |

## 6. Results and Discussion

The neurosymbolic programming approach amalgamates the advantages of neural networks and symbolic AI, resulting in a more robust model. In this study, a neural network comprising three layers (input, two hidden layers, and output layer) is initially established. The first hidden layer contains 32 units, and the second has 16 units, culminating in a single output neuron. The Rectified Linear Unit (ReLU) activation function is employed to incorporate nonlinearity into the model, as depicted in Figure 6.

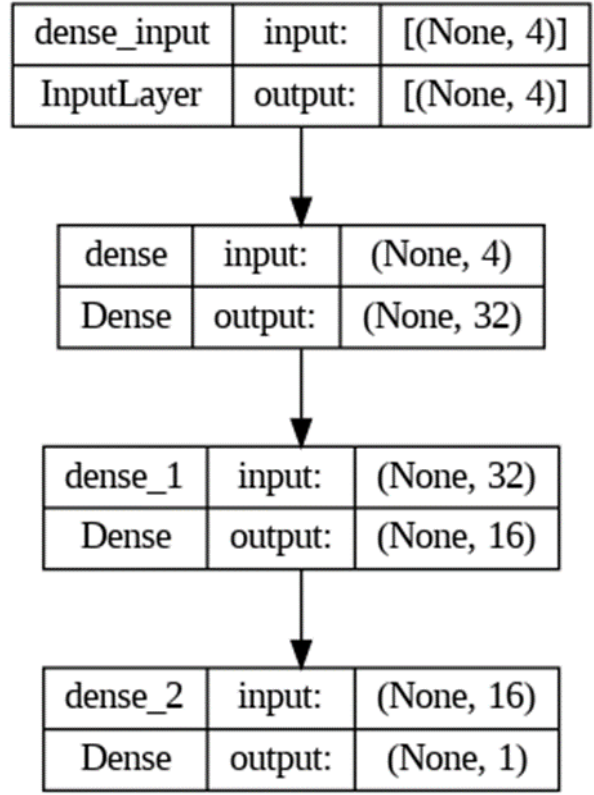

Figure 6. Neural Network architecture used in the present work

The input layer takes the input data represented by a feature vector x, where 'n' is the number of input features as shown in Equation 1. The first hidden layer applies a linear transformation using weights ($W_1$) and biases ($b_1$), followed by the ReLU activation function to introduce nonlinearity as shown in Equation 2. This results in an activation vector $a_1$ with 32 units. The second hidden layer follows a similar process using weights ($W_2$) and biases ($b_2$), resulting in an activation vector $a_2$ with 16 units as shown in Equation 3. The output layer produces the predicted output value $\hat{y}$ by applying a linear transformation using weights ($W_3$) and biases ($b_3$) as shown in Equation 4.

$x \in R^n$ (1)

$a_1 = ReLU(W_1 \cdot x + b_1)$ (2)

$a_2 = ReLU(W_2 \cdot a_1 + b_2)$ (3)

$\hat{y} = W_3 \cdot a_2 + b_3$ (4)

During the training phase, the neural network strives to minimize the mean squared error loss function illustrated in Figure 7. This function measures the disparity between the predicted output value (ŷ) and the actual output value (y), as expressed in Equation 5. The Adam optimizer updates the network's weights and biases to minimize the loss function. Once trained, the neural network can extract learned features from the input data. In this instance, the activation vector $a_2$ from the second hidden layer, containing 16 units, is utilized as the learned features (f) for the subsequent step.

$$L(y, \hat{y}) = (y - \hat{y})^2 \qquad (5)$$

Utilizing the learned features (f) and target output values (y), a decision tree regressor is trained. The decision tree serves as the symbolic component of the model, offering a human-interpretable depiction of the data relationships. To prevent overfitting, the maximum depth of the decision tree is limited to four. The trained decision tree regressor predicts output values for both the training and validation datasets using the learned features (f) as inputs. The model's performance is evaluated using the mean squared error (MSE) and R-squared ($R^2$) values. The MSE quantifies the average squared difference between the predicted and true output values, while the $R^2$ score represents the proportion of output value variance that the model can explain.

Table 2 presents the calculated MSE and $R^2$ values for both the simple ANN and the neurosymbolic approach-based algorithm.

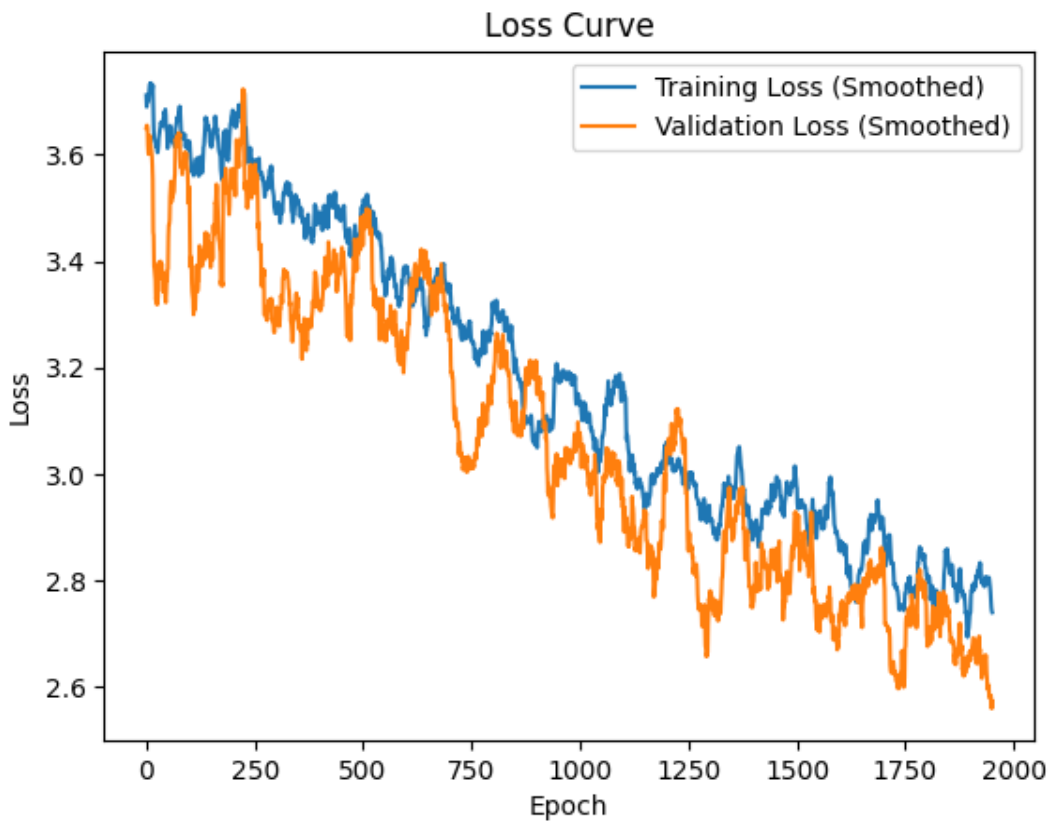

Figure 7. Decreasing loss function with the number of epochs

Table 2. Metric features evaluation of both algorithms

| Algorithm | MSE (Train) | MSE (Validation) | $R^2$ (Train) | $R^2$ (Validation) |
| --- | --- | --- | --- | --- |
| Simple ANN | 3.4174 | 3.3666 | 0.9800 | 0.9813 |
| Neurosymbolic | 2.7448 | 2.7026 | 0.9840 | 0.9850 |

Table 2 shows that the Neurosymbolic model outperforms the Simple ANN model in terms of both MSE and $R^2$ values. The MSE for the Neurosymbolic model on the training set is 2.7448, while it is 3.4174 for the Simple ANN model. Similarly, on the validation set, the Neurosymbolic model has a lower MSE (2.7026) compared to the Simple ANN model (3.3666). Lower MSE values indicate that the Neurosymbolic model has a better fit to the data, as it reduces the average squared difference between the predicted and true output values. Furthermore, the $R^2$ values of the Neurosymbolic model are higher than those of the Simple ANN model for both the training and validation sets. The $R^2$ values for the Neurosymbolic model are 0.9840 (training) and 0.9850 (validation), while the Simple ANN model has $R^2$ values of 0.9800 (training) and 0.9813 (validation). Higher $R^2$ values demonstrate that the Neurosymbolic model can explain a larger proportion of the variance in the output values compared to the Simple ANN model.

It is essential to highlight the significance of comparing true versus predicted values for both training and validation sets when evaluating a machine learning model, such as the neurosymbolic programming approach. This comparison enables researchers to measure the model's performance, assess its generalization capabilities, and ensure it is neither overfitting nor underfitting the data. The true values represent the actual, observed target outputs in the datasets, while the predicted values are generated by the model as its best estimate of the outputs based on the input features.

Comparing these values on the training set allows the assessment of the model's ability to learn the underlying patterns and relationships in the data during the training process. A good fit on the training set is crucial but not sufficient to ensure the model's effectiveness, as it may still overfit the data by memorizing noise or capturing spurious correlations. Evaluating the model on the validation set, which consists of data unseen by the model during training, provides an estimate of its generalization capabilities.

A high-performing model should maintain its accuracy and exhibit similar performance on both the training and validation sets. Discrepancies between the true and predicted values on the validation set can indicate that the model is not generalizing well to new data, potentially due to overfitting or underfitting. Figures 8 and 9 display the graphs of true versus predicted values for the training and validation sets for both the simple ANN and the neurosymbolic model.

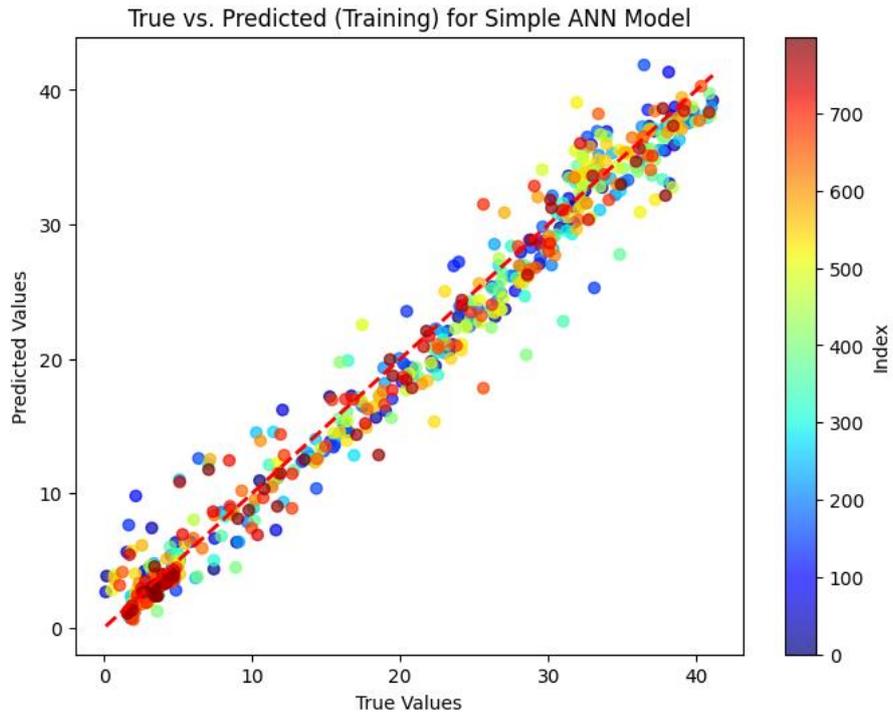

a)

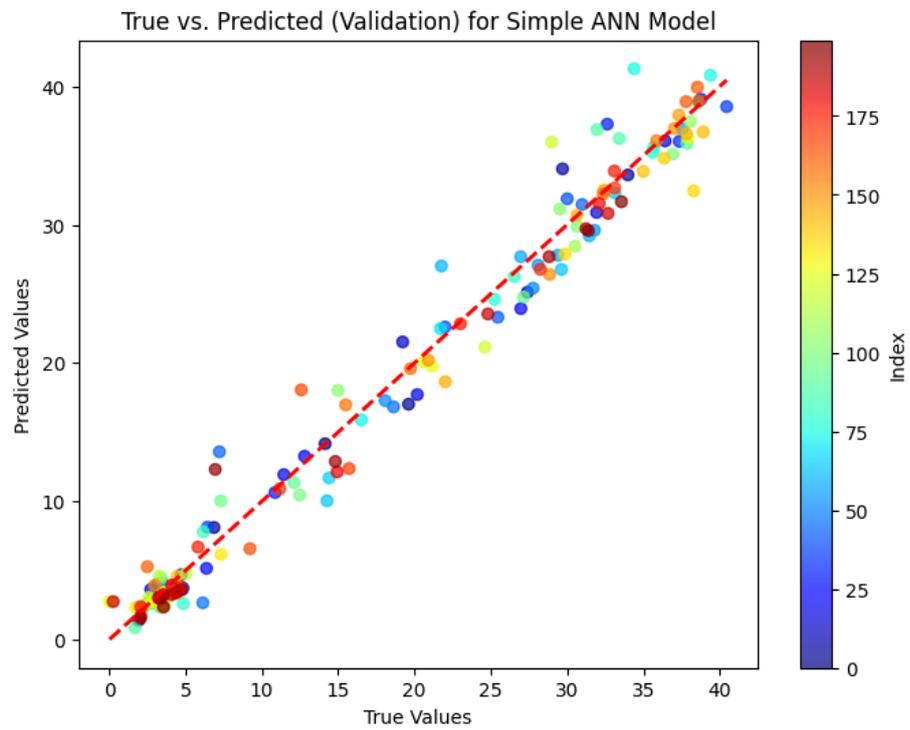

b)

Figure 8. Plot obtained in case of Simple ANN  a) true vs. predicted values for training b) true vs. predicted values for validation

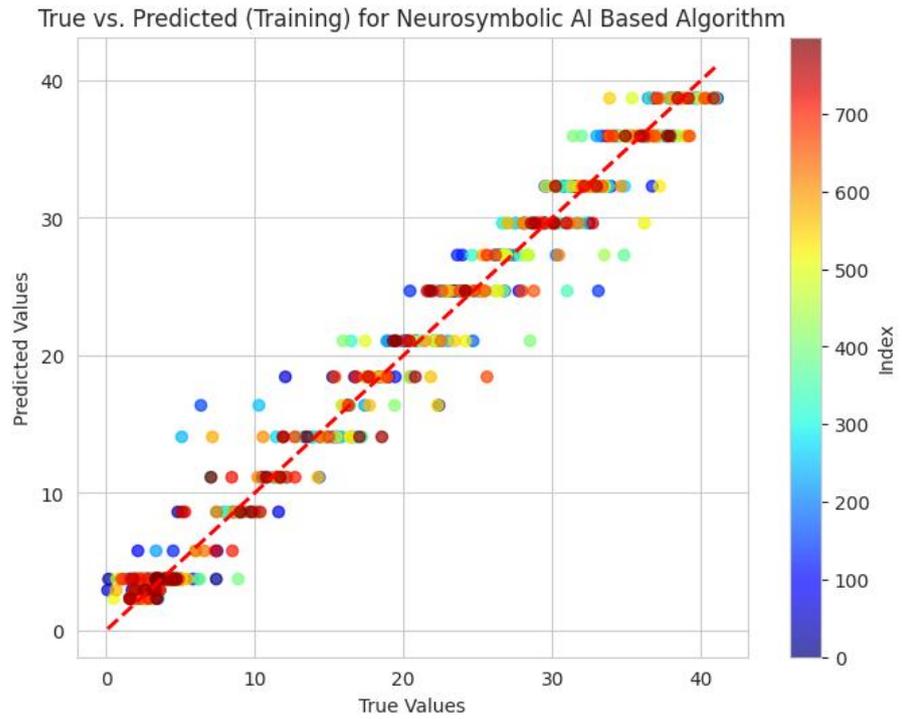

a)

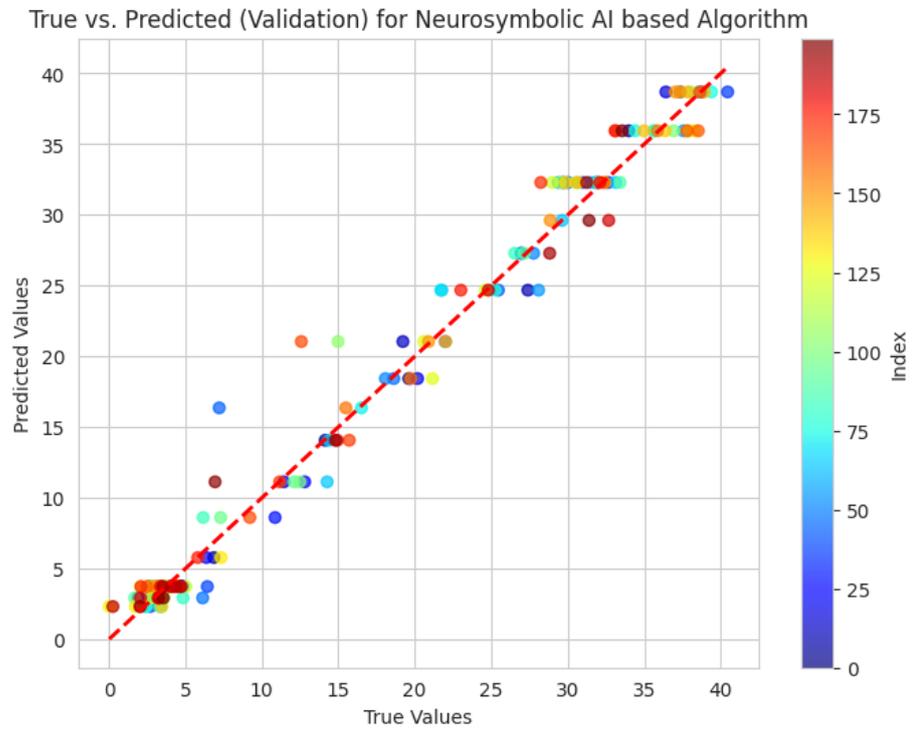

b)

Figure 9. Plot obtained in case of Neurosymbolic developed algorithm  a) true vs. predicted values for training b) true vs. predicted values for validation

## 7. Conclusion

In conclusion, this research demonstrates the effectiveness of the Neurosymbolic model in predicting the impact strength of additive manufactured polylactic acid. The model leverages the strengths of both neural networks and symbolic AI, resulting in a more robust and accurate prediction compared to the Simple ANN model. The performance metrics, including mean squared error (MSE) and R-squared ($R^2$) values, show that the Neurosymbolic model outperforms the Simple ANN model on both the training and validation sets, exhibiting superior generalization capabilities.

Future research can build upon the findings of this study by exploring several avenues. First, the Neurosymbolic model can be further fine-tuned and optimized by adjusting the neural network architecture or employing different activation functions and optimization algorithms. This may lead to even better performance in predicting the impact strength of additive manufactured materials. Second, the application of the Neurosymbolic model can be extended to other additive manufacturing materials and processes, such as metal or ceramic-based materials, and other printing techniques like selective laser sintering (SLS) or stereolithography (SLA). This will help assess the generalizability and versatility of the model across various manufacturing scenarios. Third, the integration of advanced feature selection methods and dimensionality reduction techniques can be investigated to improve the model's performance further. These approaches can help identify the most relevant input features and reduce the complexity of the model, potentially enhancing its interpretability and efficiency.